\documentclass{article}

\usepackage{amsmath}
\usepackage{arxiv}

\usepackage[utf8]{inputenc} 
\usepackage[T1]{fontenc}    
\usepackage{hyperref}       
\usepackage{url}            
\usepackage{booktabs}       
\usepackage{amsfonts}       
\usepackage{nicefrac}       
\usepackage{microtype}      
\usepackage{lipsum}
\usepackage{fancyhdr}       
\usepackage{graphicx}       
\graphicspath{{media/}}     
\usepackage{tabularx}
\usepackage{longtable}
\usepackage{multirow}
\usepackage{multicol}
\usepackage{array}
\usepackage{xcolor}
\usepackage{wrapfig} 
\usepackage{subcaption}
\usepackage{graphicx} 
\usepackage{utfsym}
\usepackage{threeparttable}
\usepackage{marvosym}
\usepackage{tcolorbox}

\usepackage{mathptmx}

\definecolor{wkyellow}{RGB}{255,241,177}
\definecolor{lightblue}{HTML}{CCE5FF}
\definecolor{lightgray}{gray}{0.9}
\definecolor{goodblue}{HTML}{0071bc}

\DeclareUnicodeCharacter{0002}{\relax}

\title{\textbf{ChineseWebText 2.0}:  Large-Scale High-quality Chinese Web Text with Multi-dimensional and fine-grained information}

\author{
Wanyue Zhang$^{1,2}$\thanks{Equal contribution, sorted by rolling dice.} \quad
Ziyong Li$^{1,2}$\footnotemark[1] \quad
Wen Yang$^{1,2}$\footnotemark[1] \quad
Chunlin Leng$^{1,2}$\footnotemark[1] \quad \\
\bf{Yinan Bai}$^{1,2}$\footnotemark[1] \quad
\bf{Qianlong Du}$^{1 *}$\footnotemark[2]  \quad
\bf{Chengqing Zong}$^{1,2}$ \quad
\bf{Jiajun Zhang}$^{1,2,3}$\thanks{Corresponding author} \quad
\\
\small{$^1$ Institute of Automation, Chinese Academy of Sciences }\quad \\
\small{$^2$ School of Artificial Intelligence, University of Chinese Academy of Sciences }\\
\small{$^3$ Wuhan AI Research }\\
\texttt{\{zhangwanyue2023, liziyong2023, yangwen2023, lengchunlin2023, baiyinan2023\}@ia.ac.cn}\\
\texttt{\{qianlong.du,cqzong,jjzhang\}@nlpr.ia.ac.cn}\\
}

\begin{document}
\date{}
\maketitle

\begin{abstract}

During the development of large language models (LLMs), pre-training data play a critical role in shaping LLMs' capabilities. In recent years several large-scale and high-quality pre-training datasets have been released to accelerate the research of LLMs, including ChineseWebText1.0~\cite{chinesewebtext}, C4~\cite{2020T5C4}, Pile~\cite{2020_pile}, WanJuan~\cite{wanjuan}, MAPCC~\cite{mapcc} and others. However, as LLMs continue to evolve, focus has increasingly shifted to domain-specific capabilities and 
safety concerns, making those previous coarse-grained texts insufficient for meeting training requirements. Furthermore, fine-grained information, such as quality, domain and toxicity, is becoming increasingly important in building powerful and reliable LLMs for various scenarios. To address these challenges, in this paper we propose a new tool-chain called \textbf{MDFG-tool} for constructing large-scale and high-quality Chinese datasets with multi-dimensional and fine-grained information. First, we employ  manually crafted rules to discard explicit noisy texts from raw contents. Second, the quality evaluation model, domain classifier, and toxicity evaluation model are well-designed to assess the remaining cleaned data respectively. Finally, we integrate these three types of fine-grained information for each text. With this approach, we release the largest, high-quality and fine-grained Chinese text \textbf{ChineseWebText2.0}, which consists of 3.8TB and each text is associated with a quality score, domain labels, a toxicity label and a toxicity score, facilitating the LLM researchers to select data based on various types of fine-grained information. The data, codes and the tool-chain are available on this website
\footnote{\url{https://github.com/CASIA-LM/ChineseWebText-2.0}}.

\end{abstract}

\section{Introduction}

In recent years, large language models(LLMs) have undergone rapid development. LLMs like BLOOM~\cite{scao2022bloom}, LLaMA~\cite{llama}, Falcon~\cite{2023refinedweb}, T5~\cite{t5}, PaLM~\cite{chowdhery2022palm}, Qwen~\cite{bai2023qwen}, GPT-4~\cite{GPT4} and O1~\cite{o1} have showcased powerful capabilities on diverse scenarios, such as question answering, mathematical problems solving, theorem proving, scientific exams, commonsense reasoning, translation, and more. During the development, a cornerstone of these LLMs' success lies in the utilization of a large-scale and high-quality dataset, which play a pivotal role in their extraordinary performance.

To facilitate research progress in LLMs, a variety of large-scale datasets have been publicly released in the last few years, such as ChineseWebText1.0~\cite{chinesewebtext},C4~\cite{2020T5C4}, Pile~\cite{2020_pile}, RefinedWeb~\cite{2023refinedweb}, WuDao~\cite{wudao}, WanJuan~\cite{wanjuan}, Yuan 1.0~\cite{yuan1}, SkyPile~\cite{skypile}, MAPCC~\cite{mapcc} and others. These datasets are typically constructed by collecting raw content from diverse sources, then applying manually designed rules or advanced classifiers to filter and refine the data, thereby producing large-scale and high-quality text corpora for LLM pre-training. However, most of these work only provide coarse-grained texts, with ChineseWebText1.0 being a notable exception by offering quality scores for each text. It should be noted that with the further development of LLMs, the focus has gradually shifted from general capabilities to domain-specific capabilities and securities. Consequently, these coarse-grained datasets are no longer sufficient to meet the growing demands of LLM training. And fine-grained annotations, such as quality, domain, toxicity and so on, are becoming increasingly crucial for boosting the domain-specific capabilities and safety of LLMs.

In order to address these issues, in this paper we propose a comprehensive tool-chain \textbf{MDFG-tool}, which can generate high-quality texts with multi-dimensional and fine-grained information from raw contents. The whole process of this approach can be divided into two stages. In the first stage, manually crafted rules, such as length filtering, sensitive word filtering, and deduplication, are applied to remove explicit noisy text from the raw data. This step produces a large-scale, high-quality dataset.
Unlike previous studies that typically end after this initial filtering, our approach incorporates a second stage comprising three advanced sub-modules: a quality evaluation model, a domain classifier, and a toxicity evaluation model. Despite the initial filtering in stage one, the diverse nature of web text often leaves a considerable amount of low-quality content. To address this, our second stage utilizes a BERT-based quality evaluation model to assign a quality score to each text, enabling researchers to curate high-quality subsets based on desired thresholds. The domain classifier, in turn, assigns domain-specific labels (e.g., technology, law, education) to each text, allowing for the efficient extraction of content tailored to specific fields, thereby supporting the development of domain-specific LLMs. Additionally, the toxicity evaluation model assigns both toxicity labels and scores, assisting researchers in enhancing the safety and ethical alignment of LLMs. Finally, these three types of fine-grained annotations are integrated for each text. Based on this complete tool-chain MDFG-tool, we release the largest Chinese dataset, ChineseWebText2.0, which features multi-dimensional and fine-grained annotations. This dataset comprises 3.8 TB of data, with each entry annotated with a quality score, domain labels, a toxicity label, and a toxicity score, empowering researchers to select data according to their specific requirements. Notably, this dataset includes a 3.16 GB subset of toxicity-labeled texts, currently the largest publicly available Chinese toxicity dataset, which can significantly enhance the toxicity evaluation capability and satefy of Chinese LLMs.

Our contributions can be summarized as follows:

(1) In this paper, we propose a new complete tool-chain MDFG-tool, which could generate high-quality Chinese pre-training data with multi-dimensional and fine-grained information.

(2) In this paper, we release the largest Chinese dataset consisting of 3.8 TB, and each text in this dataset is assigned with four types of fine-grained annotations, including quality score, domain label, toxicity label and toxicity score.

(3) In this paper, our released dataset includes a subset of 3.16 GB toxicity texts, which are currently the largest publicly available toxicity text dataset.

\section{Related Work}
\label{gen_inst}

\paragraph{Text Filtering.} During the construction of pre-training datasets, most raw data from the web contains various forms of noise, such as violence, pornography, advertisements, corrupted characters, and poorly formatted text. To extract high-quality data from these raw contents, numerous text filtering methods have been proposed to automatically remove undesirable elements. Among these methods, handcrafted rules~\cite{t5,luccioni2021whatsboxpreliminaryanalysis} are commonly used to filter out explicit noise, followed by deduplication techniques~\cite{lee2022deduplicatingtrainingdatamakes} to eliminate duplicate text from different sources. Building on this prior work, this paper introduces a rule-based preprocessing module designed to extract high-quality text from raw web content.


\paragraph{Quality Evaluation.} In addition to rule-based methods, quality evaluation approaches are also employed to identify high-quality texts using well-designed classifiers. Unlike rule-based methods, which primarily filter out explicit noise from raw content, quality evaluation approaches offer greater robustness and flexibility. These approaches leverage models such as logistic regression~\cite{brown2020language}, BERT~\cite{bert}, FastText~\cite{fasttext}, and others to calculate probability scores for each text. Based on these scores, texts are classified as positive or negative according to a predefined threshold. Among these models, BERT has emerged as one of the most widely used architectures for quality evaluation due to its exceptional performance in text classification and understanding tasks. BERT's effectiveness stems from its pre-training objectives, including masked language modeling and next-sentence prediction, which enable it to learn powerful text representation and comprehension capabilities. In this paper, we adopt the BERT architecture to build a robust quality evaluation model for text.


\paragraph{Domain Classification.} The Pre-training of LLMs on vast amounts of diverse text data often results in a generalized understanding of language. However, to enhance their effectiveness for specific applications, it is advantageous to classify and curate data from various domains, such as medicine, law, and finance. This process, known as domain classification, helps refine the model to better capture domain-specific language patterns, terminology, and contextual nuances. Similar to quality evaluation methods, domain classification approaches commonly utilize models such as logistic regression, BERT, and FastText to assign domain labels to each text. Among these, FastText is a neural network-based approach similar to CBOW~\cite{cbow}. Compared to other methods, FastText offers efficient and rapid training on large-scale datasets while maintaining comparable classification performance. As a result, this paper employs the FastText model to classify domain labels for each text.

\paragraph{Toxicity Evaluation.} 
Toxic texts are generally defined as those containing offensive or adult content~\cite{pradhan2020review, xiao2024chinese}. Offensive content can be classified into three levels of severity: offensive text, abusive text, and hate text~\cite{hatebert}. Adult content typically includes themes related to pornography or violence. Due to differences in cultural values and political perspectives, the analysis of toxic texts varies across countries. During the development of LLMs, the presence of toxic content in pre-training datasets would compromise the safety of LLMs. Consequently, toxicity evaluation has become increasingly important for the construction of pre-training datasets. In recent years, Caselli et al.~\cite{hatebert} construct a toxicity dataset, RAL-E, using data from banned Reddit communities. They then train a BERT-based toxicity evaluation model with this dataset. Hartvigsen et al.~\cite{hartvigsen2022toxigen} leverage GPT-3 to generate a dataset containing both subtle toxic texts and benign examples through adversarial techniques. In contrast, Deng et al.~\cite{deng2022cold} construct a Chinese toxicity dataset with data from Zhihu and Weibo, and annotate each text with the help of human experts and LLMs. However, previous work often suffers from small data scales and insufficient coverage, limiting its ability to effectively evaluate the toxicity of LLMs. To address the above issues, this paper aims to construct a more robust toxicity evaluation model and release a larger-scale, higher-coverage toxicity dataset, thereby enhancing toxicity evaluation capabilities and improving the safety of LLMs.

\paragraph{Datasets for Pre-training.} As the foundation of LLMs, large-scale pre-training datasets play a crucial role in enhancing the capabilities of these models. These datasets are often constructed from diverse and extensive sources, such as web crawls, books, scientific papers, and repositories like arXiv, to ensure broad coverage of knowledge. To facilitate LLM research, several large-scale datasets have been publicly released. For example, He et al.~\cite{wanjuan} introduce a 1.0 TB comprehensive dataset that includes both Chinese and English texts gathered from a wide array of online resources. Similarly, Sha Yuan et al.~\cite{wudao} develop WuDao, an ultra-large-scale Chinese corpus containing approximately 3 TB of cleaned training data extracted from 100 TB of raw web pages. This dataset spans over 50 categories, including education and technology, and utilizes over 20 cleaning rules. Chen et al.~\cite{chinesewebtext} release ChineseWebText1.0, a 1.4 TB Chinese dataset, along with a comprehensive toolchain for extracting clean web data. Notably, they introduce a quality scoring system for texts, allowing researchers to re-filter the data based on desired quality thresholds. Furthermore, IndustryCorpus~\cite{industrycorpus} presents a bilingual dataset labeled across multiple industries, combining over 100 TB of multilingual open-source datasets, such as WuDaoCorpora, RedPajama~\cite{redpajama}, and SkyPile-150B. After applying 22 domain-specific processing techniques, the resulting dataset comprises 3.4 TB of high-quality data, with 1 TB in Chinese and 2.4 TB in English. These datasets primarily focus on providing cleaned text or fine-grained information along a single dimension (e.g., quality or domain of each text). However, as LLMs continue to evolve, the focus has shifted from general capabilities to more specialized requirements, including domain-specific expertise and enhanced security considerations. This shift highlights the limitations of existing datasets, which are increasingly insufficient to meet the growing demands of training LLMs tailored for specific applications and higher ethical and security standards.

\section{Data Construction}

\begin{figure}[htp]
    \centering
    \includegraphics[width=10cm]{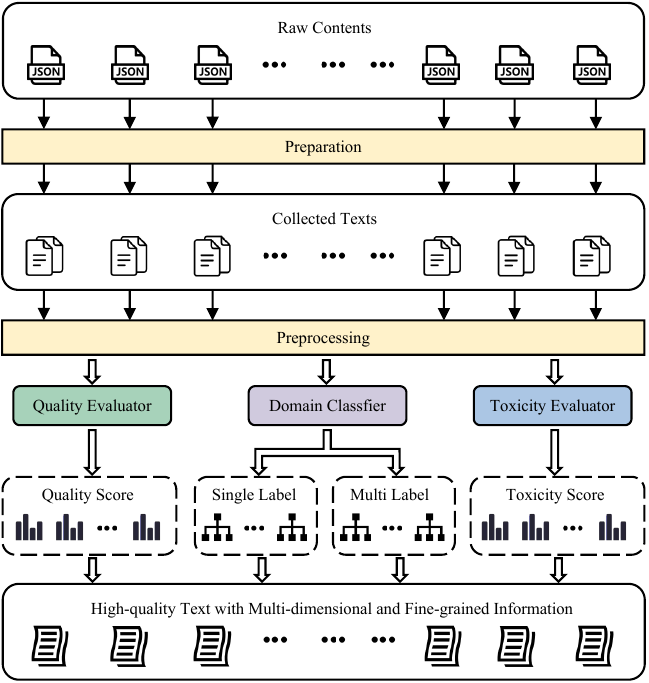}
    \caption{The pipeline of MDFG-tool.}
    \label{fig:structure}
\end{figure}

The diversity and complexity of textual data present significant challenges in constructing high-quality pre-training datasets and annotating each text with multi-dimensional, fine-grained information. To address these challenges, this paper proposes a pipeline system, MDFG-tool, which integrates a text filtering module with multi-dimensional, fine-grained annotation modules. This approach allows us to extract high-quality data from raw content and provide quality scores, domain labels, toxicity scores, and toxicity labels for each text. As shown in Figure~\ref{fig:structure}, the overview of our proposed approach is illustrated. Starting with raw content collection, we employ a preparation module to evaluate and select data from suitable sources. Next, a preprocessing module applies handcrafted rules—such as text length, character proportions, and sensitive words—to filter out noisy or irrelevant texts, yielding a high-quality dataset. Following this, three fine-grained annotation modules are applied to the refined texts: a quality evaluator assigns a quality score, a domain classifier determines the domain labels, and a toxicity evaluator calculates a toxicity score and assigns a toxicity label. Finally, these three types of annotations are integrated for each text, resulting in a high-quality dataset enriched with multi-dimensional, fine-grained metadata.

\label{headings}

\subsection{Data Collection and Preparation}
In recent years, many pre-training datasets have been publicly released, containing a variety of texts from different sources. As a result, we consider these datasets as the sources of our raw content. In this paper, we collect some publicly available datasets from the internet, including MAP-CC~\cite{mapcc}, SkyPile~\cite{skypile}, WanJuan~\cite{wanjuan}, WuDao~\cite{wudao}, and ChineseWebText1.0~\cite{chinesewebtext}, among others. All the datasets we obtained are composed of Chinese texts.


Due to the presence of noisy data and irrelevant characters in some publicly available datasets, the acquisition of high-quality data is significantly affected. To address this issue, we implement a preparation module to process data from various sources. In this module, we first sample hundreds of texts from each dataset for manual analysis. If a dataset is found to contain excessive noise or irrelevant characters, it is entirely excluded. The empirical threshold for exclusion is when the proportion of irrelevant text exceeds 30\%. Through this process, we are able to obtain the Chinese text data required for our study.




\subsection{Preprocessing}

After collecting Chinese texts from various sources, we introduce a preprocessing module designed to process all the collected texts and ensure the extraction of high-quality Chinese text data. Building upon the work of ChineseWebText1.0, this paper employs four types of handcrafted rules in the preprocessing module. The specifics of these rules are detailed below.

\textbf{Data Length.} Since most of the collected texts are from web content, a significant portion consists of short text lines that lack relevance to one another. These texts make the data unsuitable for training language models. To address this issue, following the approach of ChineseWebText1.0, we calculate the average line length for each document and discard those with an average line length of fewer than 10 characters. Additionally, short texts often provide limited information and fail to convey sufficient knowledge. Therefore, we also remove texts with fewer than 200 characters to ensure the dataset is more informative and effective for training.



\textbf{Proportion of Characters.} During the human analysis of texts in the preparation stage, we found that some Chinese datasets contain characters from other languages, non-essential characters, special symbols, and so on. These irrelevant characters are not useful for training large language models. As a result, we exclude texts that contain fewer than 30\% Chinese characters.



\textbf{Sensitive Words.} Texts from various sources often contain significant amounts of harmful content, such as references to drugs, violence, racial discrimination, and more. Such data can cause models to generate toxic content, leading to serious negative impacts on national security and social ethics. To mitigate these risks, we process all texts using a sensitive word list. Specifically, if the proportion of sensitive characters in any line of text exceeds 50\%, we classify the text as toxic and remove it from the dataset.




\textbf{Duplication.} During the data collection process, some datasets may source data from the same websites, resulting in duplicate texts across different sources. These duplicates can significantly impact the efficiency and performance of model training. To address this issue, we adopt a deduplication method to process the collected texts. Similar to ChineseWebText1.0, our approach employs a 13-gram granularity for deduplication. If the proportion of repeated characters exceeds 50\%, the duplicate text is removed from the dataset.




With the four rule-based processing methods described above, we can obtain a high-quality Chinese dataset. Most other methods would stop data processing at this stage. However, unlike these approaches, this paper introduces three evaluation models that further process the acquired high-quality data to extract multi-dimensional, fine-grained information: the quality evaluation model, domain classification model, and toxicity evaluation model.

\subsection{Quality Evaluation}

In the preprocessing phase, we applied a set of handcrafted rules to filter out explicit noisy text from our dataset. However, a significant portion of low-quality data remained, which could not be effectively eliminated by manual rules alone. To perform a more fine-grained quality assessment on the remaining data, we propose a fully automated quality evaluation method. Building on the approach used in ChineseWebText1.0~\cite{chinesewebtext}, we develop a BERT-based classification model to assign a quality score to each text. With this quality score, LLM researchers can select the desired subsets based on a specified threshold. The details of the classification model are presented below.

\begin{figure}[!h]
\centering
\begin{subfigure}[b]{0.48\textwidth}
    \includegraphics[width=\textwidth]{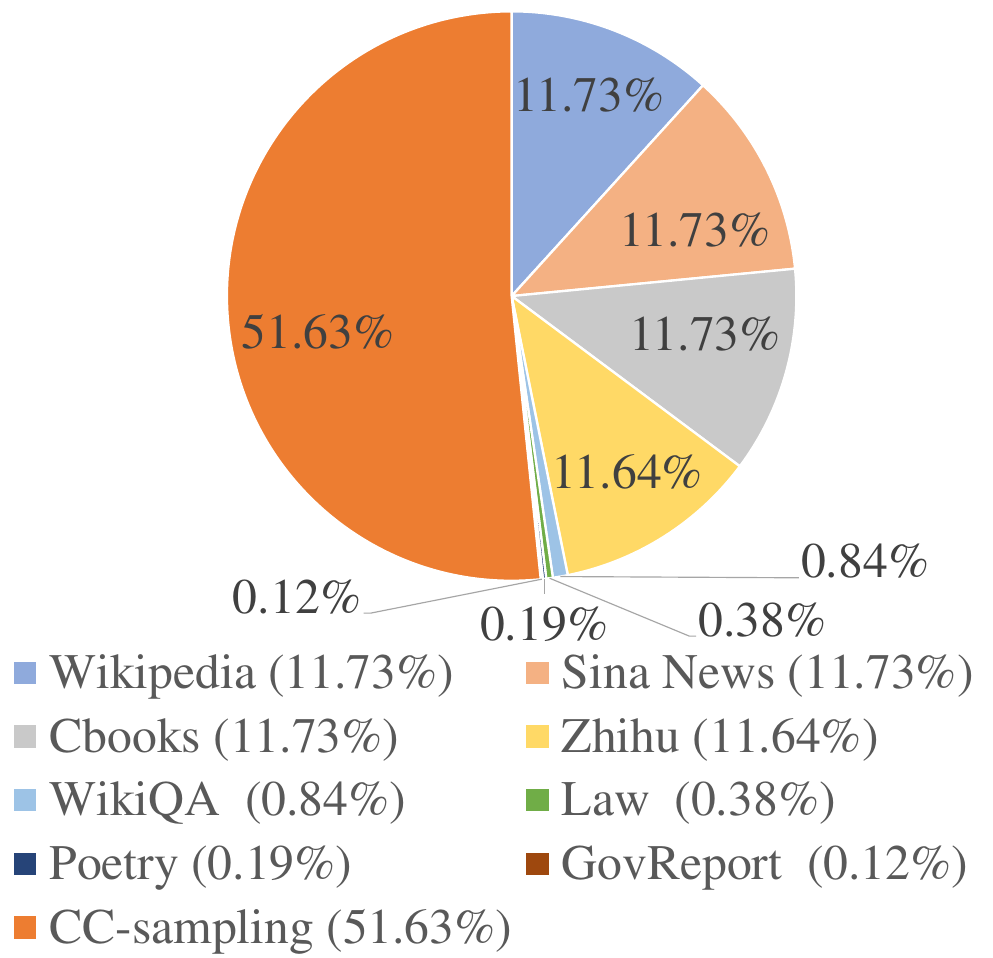}
    \caption{The Composition of BERTEval Training Data.}
    \label{fig:data_stat}
\end{subfigure}
\hfill
\begin{subfigure}[b]{0.48\textwidth} 
    \centering
    \includegraphics[width=0.6\textwidth]{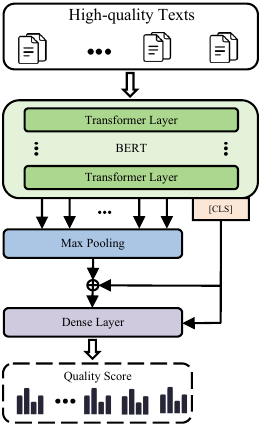} 
    \caption{The Structure of BERTEval Model.}
    \label{fig:bert_structure}
\end{subfigure}
\caption{The Overview of BERTEval Training Data Composition and Model Architecture.}
\label{fig:combined_figure_1}
\vspace{-4mm}
\end{figure}

\paragraph{Training Data}

To enhance the diversity of the training data, we incorporate a variety of text types into the positive training samples, including Wikipedia entries, e-books, poetry, news articles, and question-answer pairs. This strategy improves the model's ability to evaluate quality across a broad range of data types. Given the relatively high noise levels in Common Crawl data, we directly sample from Common Crawl to construct the negative samples. Figure~\ref{fig:data_stat} provides a detailed breakdown of the composition and quantity of the training data, demonstrating that the positive and negative samples are nearly in a 1:1 ratio.

\paragraph{Model Architecture} 

Following the work of ChineseWebText1.0, we also employ the efficient AES model Tran-BERT-MS-ML-R~\cite{wang2022use} to evaluate text quality. As shown in Figure~\ref{fig:bert_structure}, the model is based on the BERT~\cite{devlin2018bert} architecture. To reduce computational complexity, we focus solely on the text-level representation, utilizing the $[CLS]$ embedding to extract relevant information and structural features from a comprehensive perspective of the text. Simultaneously, token-level representations are derived from the sequence output of BERT, with $x$ representing the text input. After applying max pooling, the token-level representation is concatenated with the text-level representation and passed through a dense layer with Sigmoid activation, producing a text quality score $f(x|W)$ within the range (0, 1). 


\paragraph{Training Stage}

The BERT model in the ChineseWebText1.0 achieves excellent classification performance, so we utilize it as our base model. To further enhance the model's learning capability and robustness, we extend beyond the standard MSE loss~\cite{mesgar2018neural}. Specifically, we incorporate two additional loss functions: Margin Ranking (MR) loss~\cite{liu2021temp} and Cosine Similarity (CS) loss~\cite{wang2022use}. The training process consists of a pre-training phase and a self-training phase. During the pre-training phase, we train the model using positive and negative samples in a 1:1 ratio. This phase enables BERTEval to develop an initial ability to distinguish text quality.

In the Common Crawl corpus $D_n$, a significant portion of the texts meets quality standards, which may introduce inaccurate supervision and hinder BERTEval's ability to make precise quality estimations. To address this, we apply a self-training approach~\cite{scudder1965probability}. Let $S_n$ be a randomly sampled subset of $D_n$. In each iteration, BERTEval’s parameters $W^t$ from the previous step are used to generate pseudo-labels for $S_n$. BERTEval is then retrained on the sampled data in $S_n$ using the pseudo-labels to update its parameters to $W^{t + 1}$~\cite{mukherjee2020uncertainty}. Notably, as the positive samples are drawn from a large, highly reliable curated corpus that does not require pseudo-labeling, we sample only instances with pseudo-label $y_n$. 

\paragraph{Evaluation} 

       


\begin{wrapfigure}{r}{0.4\textwidth} 
\vspace{-2mm}
\centering
\includegraphics[width=0.4\textwidth]{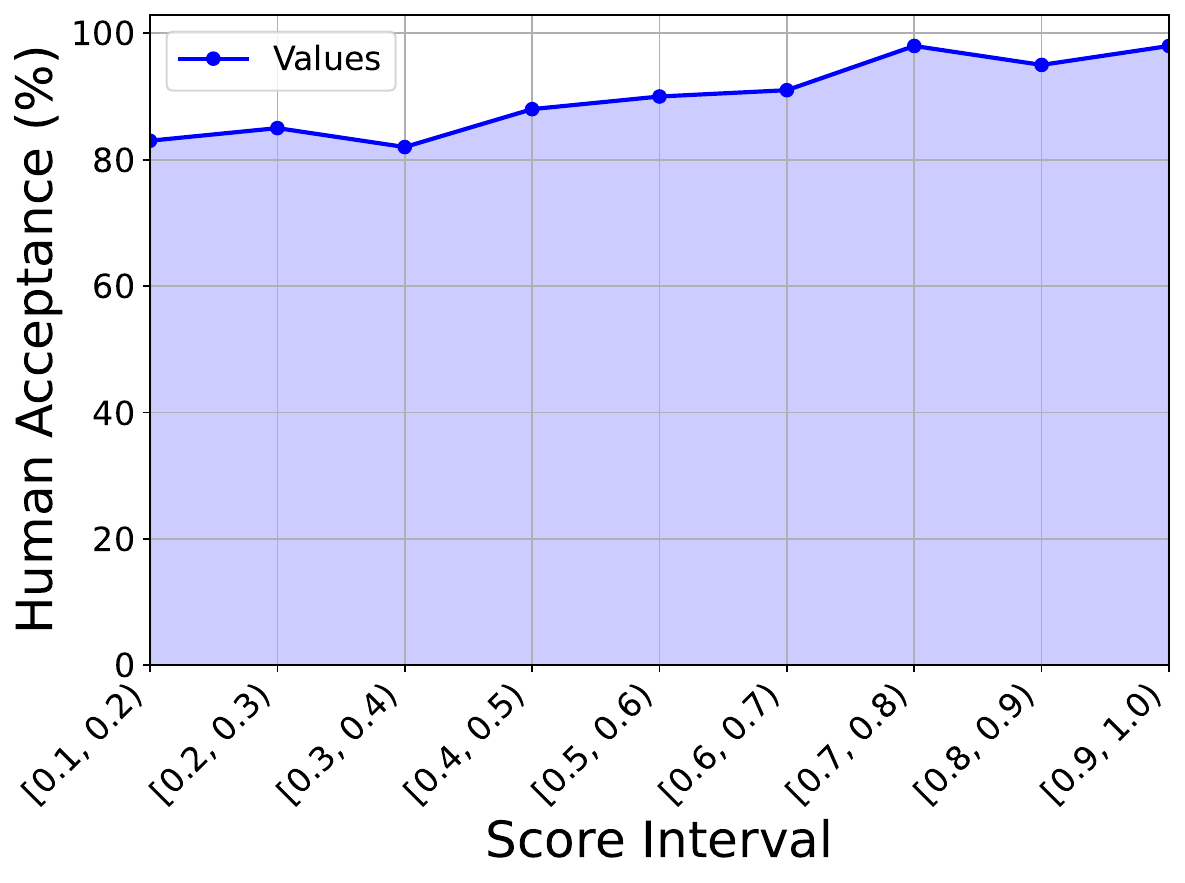}
\vspace{-4mm}
\caption{The Quality Evaluation.} 
\label{fig:human_evaluation_in_quality}
\vspace{-6mm}
\end{wrapfigure}

To evaluate the performance of BertEval in quality assessment, we conduct a human evaluation based on BertEval scores. First, we sample a validation set consisting of 50 samples from each BertEval score interval. Since the scores in the interval \([0, 0.1)\) are zero, we exclude this range from the evaluation. Then, we perform human evaluation of the quality acceptability of samples in different score intervals. The criteria for this evaluation are provided in the Appendix~\ref{appendix:human_evaluation}. Through this process, we can assess the difference between BertEval's scores and human quality acceptability.

Figure~\ref{fig:human_evaluation_in_quality} presents the results of the quality evaluation. Two key observations can be made: First, due to the high quality of our original data, human acceptability remains consistently high across the score intervals. Second, as the BertEval score increases, human acceptability shows a positive correlation, indicating strong consistency between BertEval scores and human quality assessments. This demonstrates that our quality evaluation model performs effectively in assessing content quality. 


\subsection{Domain Classification}
In the previous section, we construct a quality evaluation model for Chinese texts to facilitate the selection of high-quality training data. In contrast, this section focuses on classifying the domain of each text. In this work, we define 11 distinct domain types based on the most common application areas of LLMs: Mathematics, Books, Law, Finance, Education, Dialogue, Encyclopedia, News, Medicine, Technology, and General. We then classify each text into one of these domains using a hybrid approach that combines rule-based methods with model-driven techniques. The details of this approach are outlined below.

\subsubsection{Classification Algorithm}
\label{sec:domain classication algorithm}
\begin{figure}[!h]
\centering
\begin{subfigure}[b]{0.48\textwidth}
    \centering
    \includegraphics[width=\textwidth]{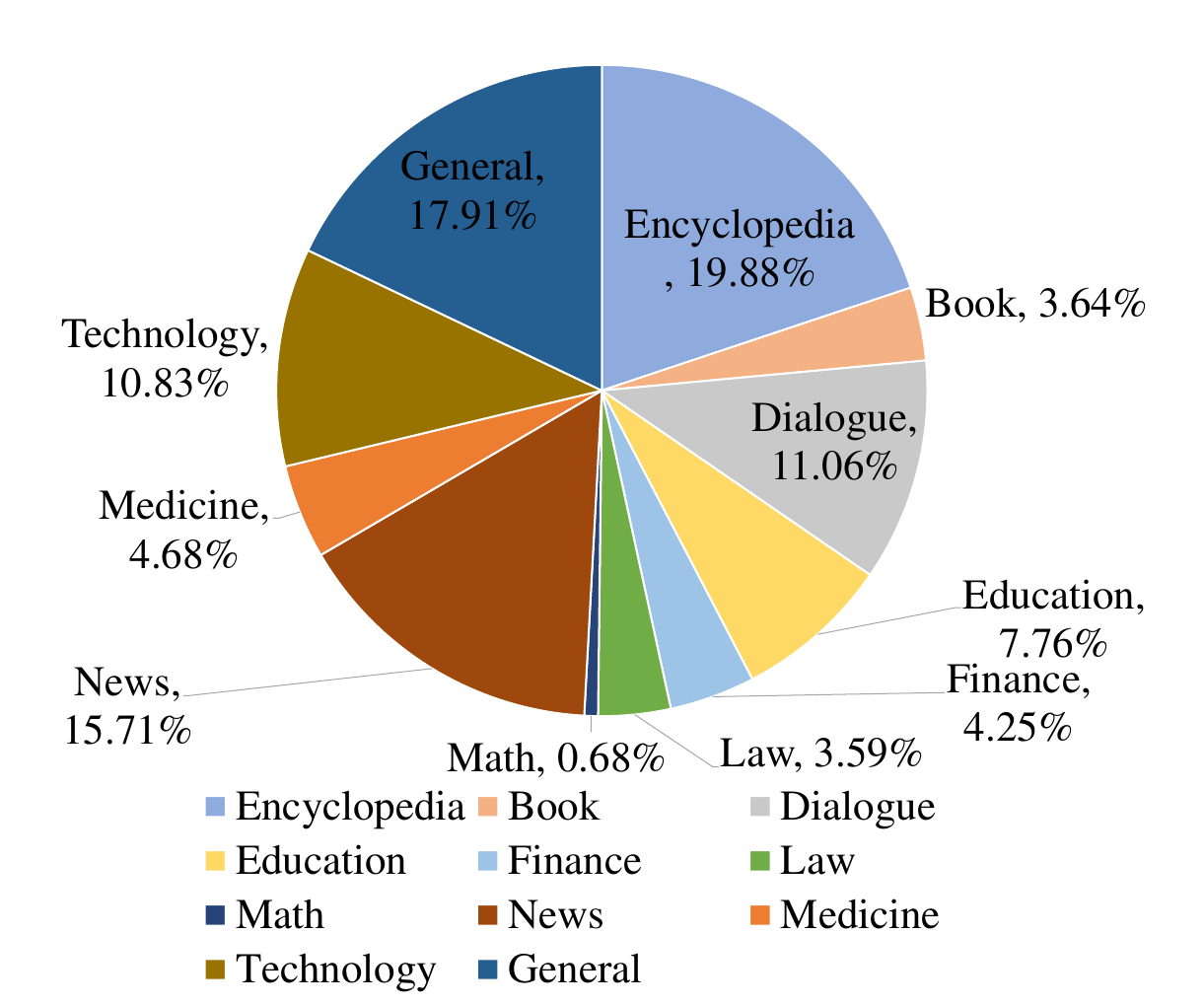}
    \caption{Distribution of Training Data across different Domains}
    \label{fig:domain_training_data}
\end{subfigure}
\hfill
\begin{subfigure}[b]{0.48\textwidth} 
    \centering
    \includegraphics[width=0.7\textwidth]{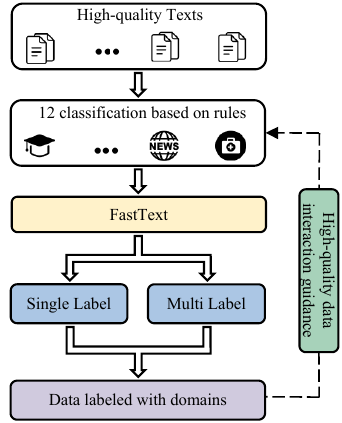}
    \caption{The Architecture of Domains Classification Algorithm}
    \label{fig:domain_structure}
\end{subfigure}
\caption{The Overview of BERTEval Training Data Composition and Model Architecture.}
\label{fig:combined_figure_2}
\vspace{-4mm}
\end{figure}

In this section, we develop a domain classification system that combines rule-based methods with a FastText-based model. FastText is a library for efficient learning of word representations and text classification. Compared to other approaches, FastText significantly reduces training and inference time while maintaining classification performance. As shown in Figure~\ref{fig:domain_structure}, the architecture of our approach is presented. In this system, the rule-based method is first used to perform initial annotation on the texts. Then, the FastText-based model is trained on the annotated data and iteratively optimized to improve performance. Notably, we annotate each text with both single and multi-labels. The details of this architecture are presented below.

\paragraph{Rule-Based Classification}
Initially, a rule-based classification approach is used to assign preliminary labels based on expert-curated keywords. For each category, 20 to 50 keywords are designated, and their occurrences in a given text are monitored, with a frequency threshold set between 3 to 5 non-repeating instances. Texts can be assigned either a single label or multiple labels based on keyword matches. If no category-specific keywords are identified, the text is labeled as "general."



\paragraph{FastText-Based Classification}
To enhance the robustness of domain classification, we introduce a FastText-based approach in addition to the rule-based methods. In this approach, the texts annotated by the rule-based method are used to guide the training of our FastText model. Specifically, two prediction strategies are implemented to support both single-label and multi-label classification. For single-label predictions, the parameter \( k=1 \) is set to return the label with the highest probability. For multi-label classification, the parameter  \( k=-1 \)  (indicating no limit on the number of returned labels) and a probability threshold of 0.3 are specified, allowing the model to return all labels with probabilities exceeding 0.3.


\paragraph{Iterative Optimization}
To further improve classification performance, we introduce an iterative optimization approach in this section. In this approach, we use a confidence threshold to filter the classification results of the FastText model and update the keyword set of the rule-based method based on these filtered results. Using the updated keyword set, we classify additional texts and integrate both the rule-based classifications and the data filtered by the confidence threshold into the training dataset. Finally, the FastText model is retrained using the updated data. Through these steps, we iteratively optimize the FastText model and construct a more powerful domain classification system.

\subsubsection{Training and Testing Data}
To enhance the diversity of the data, we collect training and testing data from various sources, including news articles, books, forums, product descriptions, and more. After several rounds of iterative optimization during the training stage, we select nearly 300,000 texts as training data and 300 texts from different domains as testing data. As shown in Table~\ref{domain_train_data}, this table presents the size of the training and testing data from different domains. Additionally, Figure~\ref{fig:domain_training_data} illustrates the detailed breakdown of the composition of the training data.



\begin{table}[!h]
	\caption{Composition of Training and Testing Data.}
	\label{domain_train_data}
	\centering
	\begin{tabular}{lcc}
		\toprule
		Domain     & Training Size ($\times 10^4 $)     &  Test Size \\
		\midrule
		book & 18581 & 20 \\ 
		dialogue & 56385 & 59 \\ 
		education & 39573 & 44 \\ 
		encyclopedia & 101393 & 107 \\ 
		finance & 21673 & 23 \\ 
		law & 18315 & 13 \\ 
		math & 3447 & 2 \\ 
		medicine & 23872 & 28 \\ 
		news & 80111 & 85 \\ 
		technology & 55237 & 56 \\ 
		general & 91353 & 89 \\ 
	    \midrule
		Total number of data & 299700 & 300 \\
		Total number of label & 509940 & 526 \\
		\bottomrule
	\end{tabular}
\end{table}


\subsubsection{Evaluation}

\paragraph{Evaluation Metric}
To rigorously evaluate the performance of the classification model, it is essential to use standard evaluation metrics that assess both its accuracy and comprehensiveness. For single-label classification task, Precision and Recall will be used, as they provide a balanced understanding of the model's performance. Precision represents the ratio of correctly predicted labels to the total number of labels predicted, while Recall is the ratio of correctly predicted labels to the total number of actual labels. The formulas for these two metrics are as follows:


\begin{equation}
\text{Precision} = \frac{\text{Number of Correctly Predicted labels}}{\text{Total Number of Predicted labels}}
\end{equation}
\begin{equation}
\text{Recall} = \frac{\text{Number of Correctly Predicted labels}}{\text{Total Number of True labels}}
\end{equation}

Different with the metric above, multi-label classification adopts micro-averaged precision and recall to evaluate the performance of models. To calculate micro-averaged precision, we sum up all true positive predictions across all labels and divide it by the sum of true positive and false positive predictions across all labels. Similarly, micro-averaged recall is computed as the total true positive predictions divided by the sum of true positive and false negative predictions. These metrics reflect the overall performance of the model across all samples and labels, treating each instance of a label equally regardless of its class. 


\paragraph{Evaluation Results}
To evaluate the classification model's performance, we manually annotated all texts in the test dataset with both single-label and multi-label classifications. As shown in Table~\ref{tab:domain_eval_manual}, the results indicate that single-label classification achieves a precision of 88.33\% and a recall of 64.15\%. For multi-label classification, we calculate the micro-averaged precision and recall, which achieve 74.48\% and 79.35\% respectively. These findings highlight the model's strong capability in domain classification, even when dealing with complex data scenarios.

\begin{table}[htbp!]
	\caption{Domain Data Evaluation.}
	\label{tab:domain_eval_manual}
	\centering
	\begin{tabular}{lcc}
		\toprule
		Label Type & Precision(\%)    &  Recall(\%) \\
		\midrule
		Single-Label & 88.33 & 64.15 \\ 
		\midrule
	 & Micro-Precision(\%)    &  Micro-Recall(\%) \\
		\midrule
		Multi-Label & 74.48 & 79.35 \\ 
		\bottomrule
	\end{tabular}
\end{table}


\subsection{Toxicity Evaluation}
Text data from the internet often contains various types of toxic content, which can significantly compromise the security of LLMs. To address this issue, we propose using a toxicity evaluation model to assess the toxicity of all texts and generate corresponding toxicity labels and scores. Due to its efficiency in reducing training and inference time while maintaining strong classification performance, FastText is once again employed to construct the toxicity evaluation model. The details of this approach are presented below.

\begin{figure}[htp]
    \centering
    \includegraphics[width=10cm]{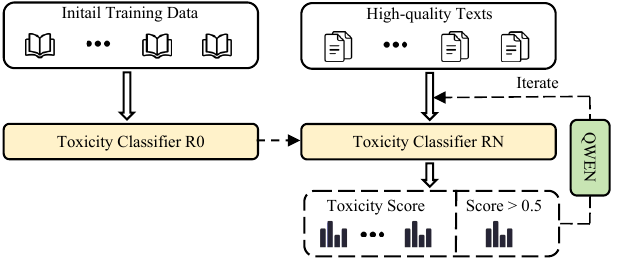}
    \caption{The Architecture of Toxicity Evaluator}
    \label{fig:Toxicity_system}
    \vspace{-4mm}
\end{figure}

\subsubsection{Methodology}
In the training process of toxicity evaluation models, the scale and quality of training data have a significant impact on the performance of the models. However, an analysis of existing datasets reveals that toxicity data is often limited in size and lacks diversity, which substantially restricts the effectiveness of these models. At the same time, testing publicly available large language models (LLMs) has shown that some of these models possess strong toxicity evaluation capabilities, enabling them to annotate toxicity data both quickly and accurately. Based on this observation, this paper proposes an LLM-in-the-loop approach to improve the training of toxicity evaluation models, resulting in significant performance enhancements. The details of this method are outlined below.




\paragraph{Initail Training}
In the training process of evaluation models, we first combine several high-quality Chinese toxicity datasets—COLD~\cite{deng2022cold}, ToxiCN~\cite{toxicn}, SWSR~\cite{jiang2022swsr}, and CDial-Bias~\cite{zhou2022towards}—to create an initial training set. To increase the diversity of the dataset and better align its distribution with that of our specific dataset, we sample a subset of texts from our high-quality dataset to serve as benign samples. Additionally, to ensure a balanced distribution between toxic and benign samples, we doubled the number of toxic samples in the initial training set. With this augmented dataset, we train the initial FastText model, referred to as Toxic Classifier R0. As shown in Table~\ref{toxicity_data}, it presents a detailed breakdown of the source and scale of each portion of the initial training dataset.

\begin{table}[htbp]
	\caption{Composition of Toxicity Initial Training Data.}
	\label{toxicity_data}
	\centering
	\begin{tabular}{llc}
		\toprule
		Source     & Toxic Size     &  Benign Size \\
		\midrule
		COLD & 36162 & 19439 \\ 
		ToxiCN & 12922 & 5550 \\ 
		SWSR & 6186 & 5876 \\ 
		CDial-Bias & 12986 & 19013 \\ 
		Samples & 0 & 80000 \\ 
	    \midrule
		Sum & 68256 & 129878 \\
		\bottomrule
	\end{tabular}
\end{table}

\paragraph{LLM-in-the-loop Training}
In this section, LLMs will be utilized to further enhance the performance of toxicity evaluation models. In this approach, we first select a subset from our large-scale dataset and then Toxic Classifier R0 is employed to score each sample. Next, texts with toxicity scores greater than 0.5 will be screened out to form a candidate set. After that the candidate subset will be scored by a LLM, which categorize them into toxic and benign classes, thereby forming a new training set. Here, the chosen LLM is Qwen2.5-32B-Instruct~\cite{qwen2.5}. Finally, we integrate the new training set with the initial training set, and then retrain the FastText model with this refined training dataset. In this paper, we conduct two iterations on the above method, significantly improving the performance of the evaluation model.

\subsubsection{Evaluation}
To conduct a comprehensive evaluation across data from various sources, a test set is randomly selected from the large-scale dataset and manually annotated with toxic or benign labels. Due to the imbalance between toxic and benign samples in our dataset, toxic and benign samples are collected separately during the test data sampling. The final test set consists of 300 toxic samples and 300 benign samples.


Given the sampling methodology of the test set, Precision and Specificity were selected as metrics to evaluate the performance of the model on toxic and benign data, respectively. In this study, toxicity is defined as the positive class, whereas benignity is defined as the negative class. Precision is defined as the ratio of true positive predictions to total positive predictions, while Specificity measures the ratio of true negative predictions to total negative predictions. The formulas of these two metrics are shown below.

\begin{equation}
\text{Precision} = \frac{\text{True Positive}}{\text{True Positive + False Positive}}
\end{equation}
\begin{equation}
\text{Specificity} = \frac{\text{True Negative}}{\text{True Negative + False Negative}}
\end{equation}

As shown in Table~\ref{toxicity_evaluation}, the evaluation results of our model are presented. From this table, we can observe that the precision of our toxicity evaluation model for toxic texts reaches 83.67\%, while the specificity for benign texts is 97.67\%. These results indicate that our model demonstrates excellent performance in toxicity evaluation.

\begin{table}[htbp]
	\caption{The Evaluation of Toxicity Classifier.}\label{toxicity_evaluation}
	\centering
	\begin{tabular}{lccc}
		\toprule
		Category     & Precision ($\%$)  & TP+FP   &  TP \\
		\midrule
		Toxic & 83.67 & 300 & 251   \\ 
		\midrule
		             & Specificity ($\%$)  & TN+FN   &  TN \\
		\midrule
		Benign & 97.67 & 300 & 293   \\ 
		\bottomrule
	\end{tabular}
\end{table}

\subsection{Dataset Statistics and Comparison}
After processing the collected data with the modules outlined above, this paper constructs a clean Chinese text dataset, ChineseWebText2.0, which consists of 3.8 TB of data. Each text in this dataset is assigned a quality score, domain labels, a toxicity score, and a toxicity label. As shown in Table~\ref{Comparison}, we compare our dataset with several other publicly available pre-training corpora. Most of these datasets primarily focus on collecting clean texts from various sources, while neglecting fine-grained text annotations. In this comparison, C4~\cite{2020T5C4}, The Pile~\cite{2020_pile}, and REFINEDWEB~\cite{2023refinedweb} are three public English datasets, while WuDaoCorpora~\cite{wudao}, ROOTS-zh~\cite{roots}, WanJuan1.0-zh~\cite{wanjuan}, MAPCC~\cite{mapcc}, IndustryCorpus~\cite{industrycorpus}, and ChineseWebText1.0~\cite{chinesewebtext} are Chinese corpora. Specifically, the texts in IndustryCorpus are tagged with domain labels, while each text in ChineseWebText1.0 is annotated with a quality score. Compared to these prior works, our dataset is both the largest and most recent Chinese dataset, and it features four types of fine-grained annotations for each text. This extensive annotation can aid LLM researchers in enhancing model performance in specific domains while improving safety.

\begin{table}[htbp!]
\begin{center} 
    \begin{threeparttable}
        \caption{Overview of output datasets. Note that QS$=$Quality Scoring, DC$=$Domain Classification, TC$=$Toxicity Classification, TS$=$Toxicity Scoring.}
        \label{Comparison}
        \begin{tabular}{*{8}{c}}
            \toprule 
            \multirow{2}{*}{Dataset} & \multirow{2}{*}{Lang.} & \multirow{2}{*}{Availability} & \multirow{2}{*}{Public Size} & \multicolumn{4}{c}{Fine-grained Information}\\
            
            \cmidrule(lr){5-8}  & & & & QS & DC & TC & TS\\
            \midrule
            C4~\cite{2020T5C4} & EN & Public & 807GB & \textcolor{red}{\usym{1F5F4}} & \textcolor{red}{\usym{1F5F4}} & \textcolor{red}{\usym{1F5F4}} & \textcolor{red}{\usym{1F5F4}}\\
            The Pile~\cite{2020_pile} & EN & Public & 825GB & \textcolor{red}{\usym{1F5F4}} & \textcolor{red}{\usym{1F5F4}} & \textcolor{red}{\usym{1F5F4}} & \textcolor{red}{\usym{1F5F4}}\\
            REFINEDWEB~\cite{2023refinedweb} & EN & Public & 2.8TB & \textcolor{red}{\usym{1F5F4}} & \textcolor{red}{\usym{1F5F4}} & \textcolor{red}{\usym{1F5F4}} & \textcolor{red}{\usym{1F5F4}}\\
            WuDaoCorporal~\cite{wudao} & ZH & Pratly Public & 200GB & \textcolor{red}{\usym{1F5F4}} & \textcolor{red}{\usym{1F5F4}} & \textcolor{red}{\usym{1F5F4}} & \textcolor{red}{\usym{1F5F4}}\\
            ROOTS-zh~\cite{roots} & ZH & Public & 265GB & \textcolor{red}{\usym{1F5F4}} & \textcolor{red}{\usym{1F5F4}} & \textcolor{red}{\usym{1F5F4}} & \textcolor{red}{\usym{1F5F4}}\\
            WanJuan1.0-zh~\cite{wanjuan} & ZH & Public & 550GB & \textcolor{red}{\usym{1F5F4}} & \textcolor{red}{\usym{1F5F4}} & \textcolor{red}{\usym{1F5F4}} & \textcolor{red}{\usym{1F5F4}}\\
            MAPCC~\cite{mapcc} & ZH & Public & 3TB & \textcolor{red}{\usym{1F5F4}} & \textcolor{red}{\usym{1F5F4}} & \textcolor{red}{\usym{1F5F4}} & \textcolor{red}{\usym{1F5F4}}\\
            IndustryCorpus-zh~\cite{industrycorpus} & ZH & Public & 1TB & \textcolor{red}{\usym{1F5F4}} & \textcolor{green}{\usym{2714}} & \textcolor{red}{\usym{1F5F4}} &\textcolor{red}{\usym{1F5F4}}\\
            ChineseWebText1.0~\cite{chinesewebtext} & ZH & Public & 1.4TB & \textcolor{green}{\usym{2714}} & \textcolor{red}{\usym{1F5F4}} & \textcolor{red}{\usym{1F5F4}} & \textcolor{red}{\usym{1F5F4}}\\
            \textbf{ChineseWebText2.0} & ZH & Public & 3.8TB & \textcolor{green}{\usym{2714}} & \textcolor{green}{\usym{2714}} & \textcolor{green}{\usym{2714}} & \textcolor{green}{\usym{2714}}\\ 
            \bottomrule
        \end{tabular}
    \end{threeparttable}
\end{center}
\end{table}

\section{Data Analysis}
\subsection{Removal Rate for Different Stages}
\begin{figure}[htpb!]
    \centering
    \includegraphics[width=0.8\textwidth]{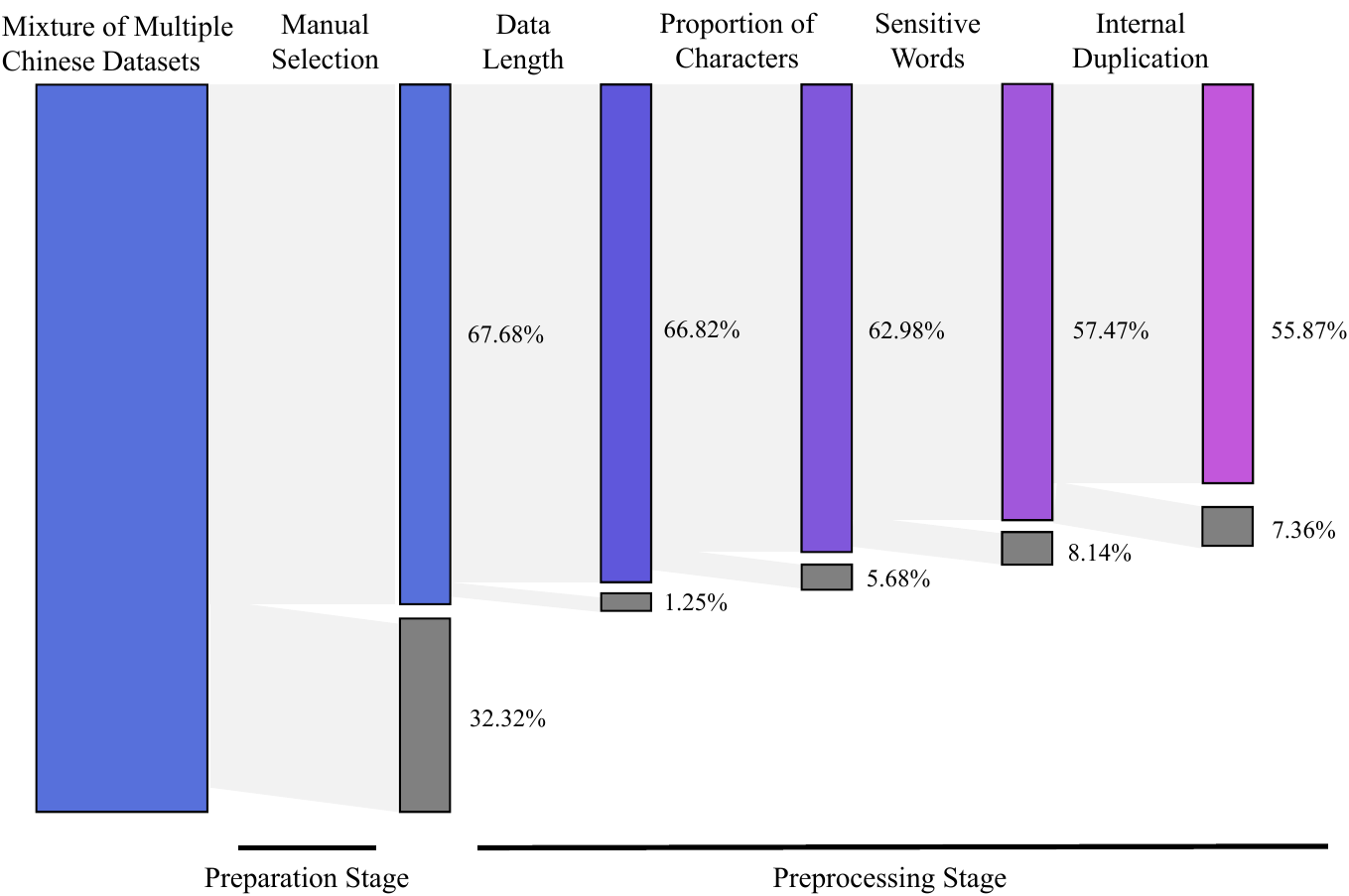}
    \caption{The proportion of data removed from the originally collected data in each processing step. The gray bars represent the proportion of data removed in each step relative to the data remaining before that step, while the other colored bars represent the retained data and its proportion relative to the originally collected data.}
    \label{fig:fig4removal-rate}
\end{figure}

To provide a high-level overview of the preparation and preprocessing stages, Figure~\ref{fig:fig4removal-rate} illustrates the processing workflow along with the removal rate at each step. This figure shows the proportion of data removed at each stage and the percentage of remaining data relative to the original collected dataset. This helps readers track the progression of data through the various processing stages, from raw data to the final high-quality dataset.


After collecting raw data from various sources, we initially obtain an original Chinese dataset totaling 6.6 TB. However, due to a significant amount of irrelevant and noisy content in some sources, a manual sampling analysis is performed during the preparation stage. If irrelevant text accounted for more than 30\% of a source, that data is discarded entirely. As a result, a substantial portion of the data is removed, leaving 67.68\% of the original dataset. In the preprocessing stage, four rule-based steps are implemented to filter the remaining data. First, the Data Length step remove overly short texts to ensure each text contained sufficient informational content. Next, the Character Proportion step eliminate texts with a high percentage of noisy characters, such as English, Traditional Chinese characters, or other irrelevant symbols. Finally, the Sensitive Words and Deduplication steps are applied to remove toxic content and duplicate texts from the dataset. After the preprocessing stage, a high-quality Chinese text dataset totaling 3.8 TB is produced. In the next stage, each text in this high-quality dataset will be enriched with fine-grained annotations, including a quality score, domain labels, a toxicity score, and a toxicity label.

\subsection{Data Quality Distribution}

\begin{figure}[!h]
\centering
\begin{subfigure}[b]{0.48\textwidth}
    \includegraphics[width=\textwidth]{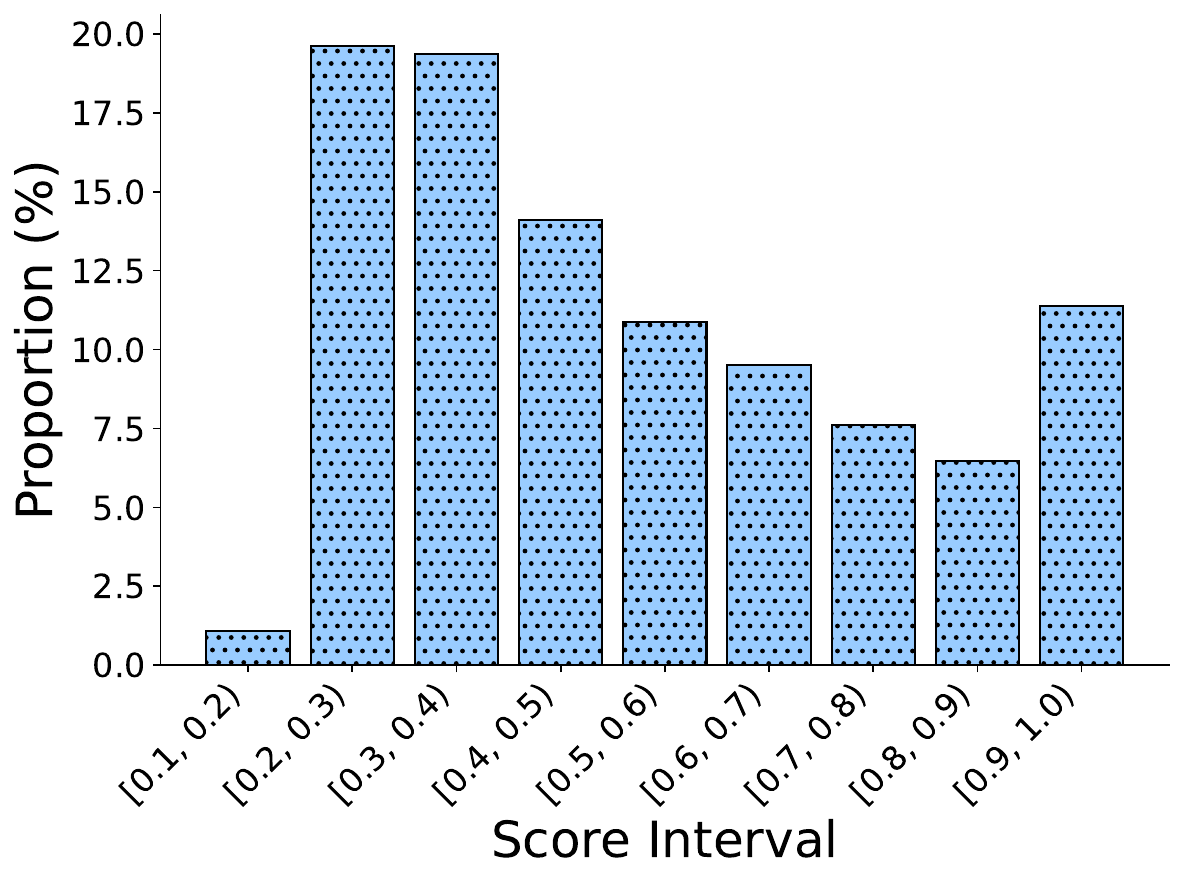}
    \caption{Quality Distribution}
    \label{fig:quality_distribution}
\end{subfigure}
\hfill
\begin{subfigure}[b]{0.48\textwidth} 
    \includegraphics[width=\textwidth]{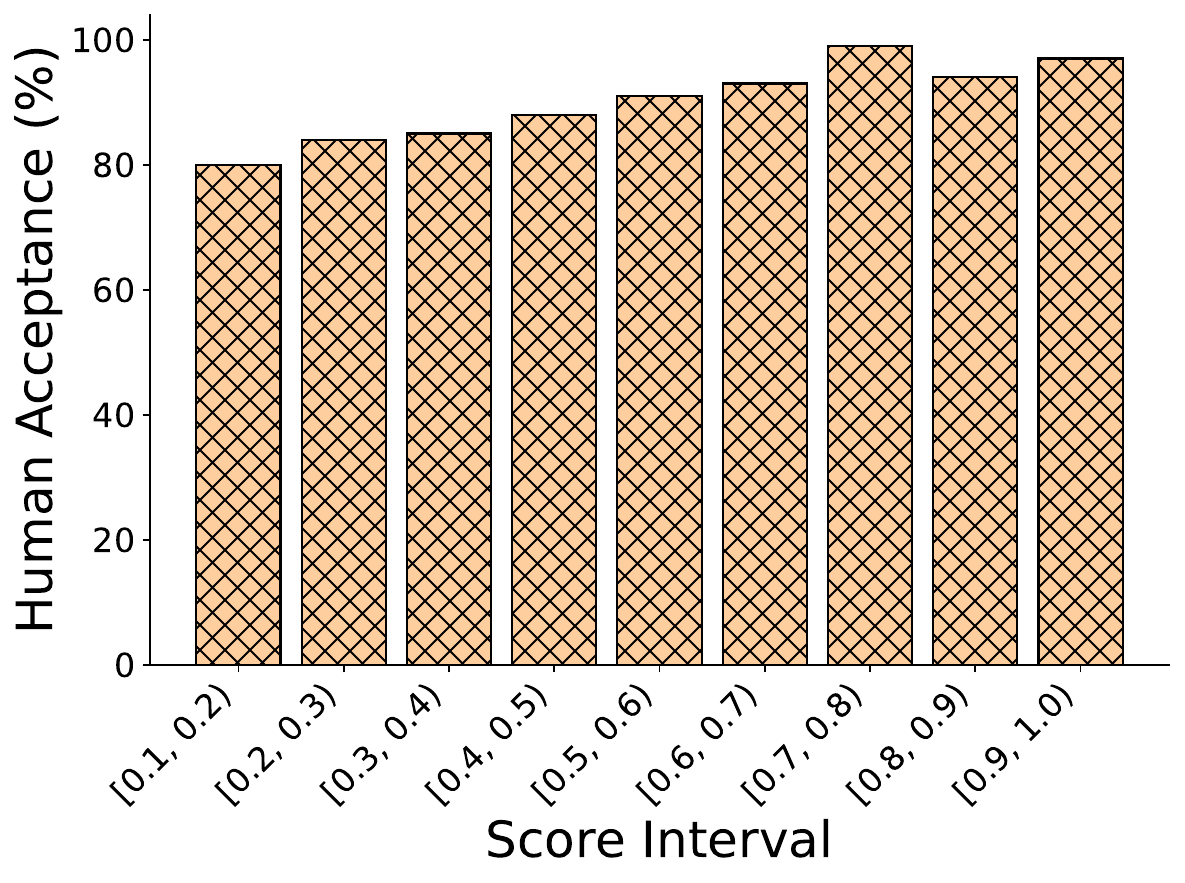}
    \caption{Human Acceptance Evaluation}
    \label{fig:human_acceptance}
\end{subfigure}
\caption{The Data Analysis on Quality Evaluation.}
\label{fig:combined_figure_3}
\vspace{-6mm}
\end{figure}

\paragraph{Quality Distribution} To investigate the quality distribution, we calculate the data proportions across different quality score ranges from our ChineseWebText 2.0 dataset. Figure~\ref{fig:quality_distribution} shows the proportion of data across different quality score intervals. The data is primarily concentrated in the mid-range score intervals \([0.2, 0.4)\), each contributing approximately 18\%. Additionally, a significant proportion lies within the high-quality interval \([0.9, 1.0)\), reflecting the presence of high-quality content in the dataset. In contrast, the lowest interval \([0.1, 0.2)\)  contains only a minimal fraction, indicating a limited amount of poor-quality data. Note that the quantity of quality scores in the range [0, 0.1) is zero, so this interval has been omitted. This quality distribution provides a valuable reference for LLM researchers, enabling them to select data based on desired quality thresholds.

\paragraph{Human Acceptance Evaluation}
To validate the consistency between quality evaluation and human judgments, we randomly sample 100 samples from each score interval. The details and criteria for the human evaluation are provided in the Appendix~\ref{appendix:human_evaluation}. Figure~\ref{fig:human_acceptance} displays human acceptance rates across different score intervals, showing a clear positive trend: higher scores correlate with higher acceptance rates. 
Specifically, the highest score interval \([0.5, 1.0)\) achieves an acceptance rate exceeding 90\%, while the lowest interval \([0.1, 0.2)\) still maintains an acceptance rate of 80\%. This trend highlights the overall high quality of the data. 

In summary, the dataset is primarily concentrated in the mid-quality range, with higher scores strongly correlating to greater human acceptance. This alignment highlights the dataset's potential for high-quality applications, where consistency in human-like quality is crucial.

\subsection{Domain Distribution}

To investigate the distribution of our dataset across different domains, in this section, we conduct an detailed analysis of the data distribution across eleven distinct domains: \textit{book}, \textit{dialogue}, \textit{education}, \textit{encyclopedia}, \textit{finance}, \textit{law}, \textit{math}, \textit{medicine}, \textit{news}, \textit{technology}, and \textit{general}. This analysis considers two perspectives: the overall domain distribution and the quality-related domain distribution, providing comprehensive insights into the dataset's composition across different domains.


\begin{figure}[htpb!]
    \centering
    \begin{minipage}{0.45\textwidth}
        \centering
        \begin{tabular}{lcc}
            \toprule
            Domain & Sample Count & Proportion (\%) \\
            \midrule
            Book & 45,192,607 & 2.97 \\
            Dialogue & 312,324,923 & 20.54 \\
            Education & 219,530,073 & 14.44 \\
            Encyclopedia & 508,376,352 & 33.43 \\
            Finance & 110,326,424 & 7.26 \\
            Law & 40,477,319 & 2.66 \\
            Math & 8,331,242 & 0.55 \\
            Medicine & 104,809,588 & 6.89 \\
            News & 425,848,698 & 28.01 \\
            Technology & 295,978,428 & 19.47 \\
            General & 508,392,319 & 33.44 \\
            \bottomrule
            \\
            \\
        \end{tabular}
        \caption{Data Distribution Across Domains}
        \label{tab:domain_distribution}
    \end{minipage}
    \hfill
    \begin{minipage}{0.50\textwidth}
        \centering
        \includegraphics[width=\linewidth]{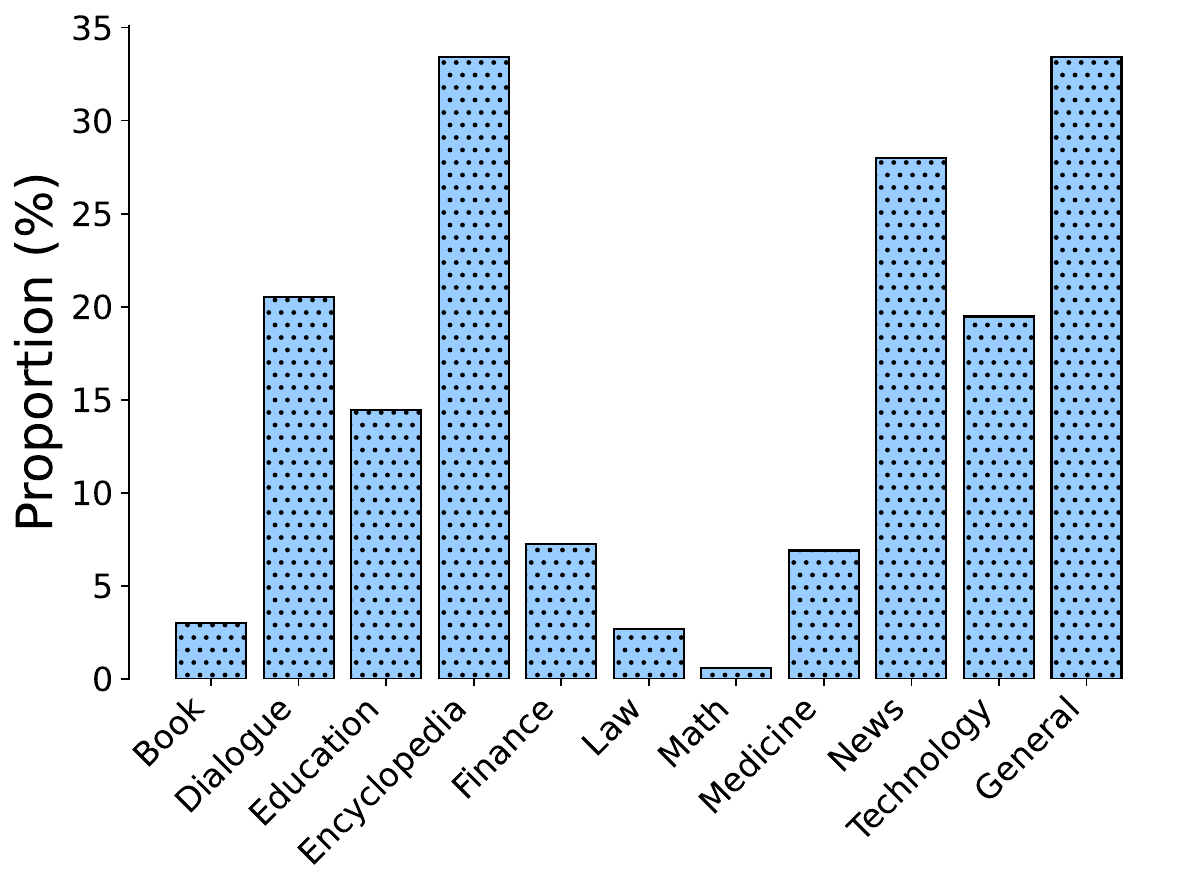}
        \caption{Data Distribution Across Domains.}
        \label{fig:domain_distribution}
    \end{minipage}
    \vspace{-6mm}
\end{figure}

\paragraph{Overall Domain Distribution}
As shown in Figure~\ref{tab:domain_distribution}, the sample counts and corresponding proportions across various domains are presented. The Encyclopedia, General, and News domains dominate the dataset, comprising 33.43\%, 33.44\%, and 28.01\% of the total data, respectively. In contrast, the Math domain has the smallest share at 0.55\%, yet it still includes over 8 million samples. Figure~\ref{fig:domain_distribution} provides a bar chart that offers a more intuitive visualization of this distribution. This comprehensive domain analysis allows LLM researchers to select datasets that enhance the model's knowledge and capabilities in specific areas.

\paragraph{Quality-Related Domain Distribution}
In order to explore the domain distribution across different quality intervals, we perform an analysis focused on the quality-related domain distribution. Specifically, we calculate the proportions of various domains within each quality interval. As shown in Table~\ref{tab:quality_domain_distribution}, this table provides a detailed breakdown of domain proportions across different quality intervals. From the results, we observe that the distribution of domain data within each quality interval aligns closely with their overall distribution in the dataset. Based on the proportions in Table~\ref{tab:quality_domain_distribution}, researchers can select domain-specific data within targeted quality intervals, enabling the extraction of higher-quality domain-specific data subsets.


\begin{table}[htpb!]
    \centering
    \caption{Domain Distribution Across Quality Levels.}
    \label{tab:quality_domain_distribution}
    \begin{tabular}{lcccccccccc}
        \toprule
        \multirow{2.5}{*}{Domain} & \multicolumn{9}{c}{Proportion in Each Quality Interval (\%)} \\
        \cmidrule{2-11}
        & 0.1-0.2 & 0.2-0.3 & 0.3-0.4 & 0.4-0.5 & 0.5-0.6 & 0.6-0.7 & 0.7-0.8 & 0.8-0.9 & 0.9-1.0 & \textbf{Total} \\
        \midrule
        Book & 2.37 & 1.98 & 2.61 & 2.72 & 2.57 & 2.83 & 3.27 & 3.83 & 5.47 & 2.97 \\
        Dialogue & 7.99 & 12.57 & 17.48 & 21.57 & 24.99 & 26.32 & 27.79 & 27.94 & 21.26 & 20.54 \\
        Education & 8.43 & 11.70 & 15.00 & 15.33 & 15.96 & 16.11 & 15.44 & 14.82 & 13.93 & 14.44 \\
        Encyclopedia & 17.98 & 27.67 & 31.51 & 34.92 & 38.66 & 40.03 & 39.52 & 37.29 & 29.48 & 33.43 \\
        Finance & 1.57 & 3.16 & 4.82 & 6.67 & 8.81 & 10.48 & 11.86 & 12.23 & 9.65 & 7.26 \\
        Law & 0.34 & 0.80 & 1.43 & 1.97 & 2.63 & 3.54 & 4.71 & 5.46 & 5.40 & 2.66 \\
        Math & 0.42 & 0.41 & 0.48 & 0.58 & 0.65 & 0.68 & 0.71 & 0.67 & 0.49 & 0.55 \\
        Medicine & 3.21 & 6.72 & 7.71 & 7.02 & 7.25 & 7.28 & 7.00 & 6.61 & 5.41 & 6.89 \\
        News & 18.78 & 19.60 & 24.33 & 28.55 & 32.62 & 34.78 & 34.72 & 33.61 & 31.19 & 28.01 \\
        Technology & 16.75 & 18.52 & 18.44 & 20.42 & 22.26 & 22.42 & 21.45 & 19.66 & 15.32 & 19.47 \\
        General & 53.56 & 44.24 & 35.12 & 31.18 & 27.47 & 25.62 & 25.36 & 26.69 & 32.61 & 33.44 \\
        \bottomrule
    \end{tabular}
\end{table}


\subsection{Data Toxicity Analysis}

\begin{figure}[htp]
    \centering
    \includegraphics[width=15cm]{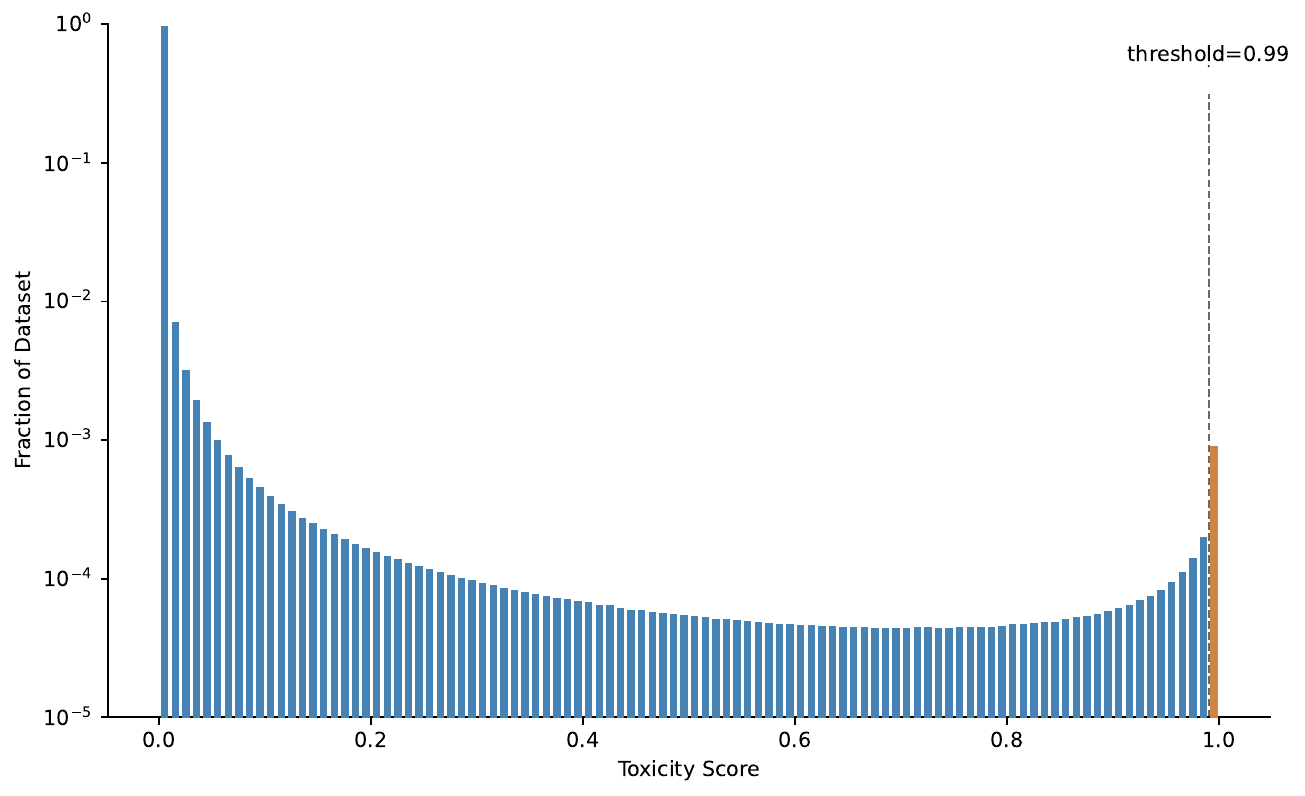}
    \caption{The Distribution of Toxicity: A threshold of 0.99 was established, and samples with scores exceeding 0.99 were categorised as toxic.}
    \label{fig:Toxicity_distribution}
\end{figure}

During the training procedure of LLMs, toxic data introduces harmful knowledge, which can lead the model to generate toxic outputs. In this section, we analyze the toxicity distribution within our dataset. As shown in Figure~\ref{fig:Toxicity_distribution}, the figure illustrates the toxicity distribution of the dataset, where a higher toxicity score indicates greater toxicity. It is evident that the majority of the data in our dataset has a toxicity score of 0.0, signifying non-toxic, high-quality data. These non-toxic texts make up 97.41\% of the dataset.

Additionally, through manual analysis of the toxicity scores, we identify that data with scores above 0.99 are classified as toxic. By applying this empirical threshold, we filter our dataset to obtain a 3.16GB subset of toxic texts, comprising 1,632,620 samples. In Table~\ref{tab:comparison_toxic}, we compare this subset with other publicly available toxic datasets. The table includes OffensEval 2019~\cite{offenseval}, AbusEval~\cite{abuseval}, HatEval~\cite{hateval}, RAL-E~\cite{hatebert}, and ToxiGen~\cite{hartvigsen2022toxigen}, which are English toxicity datasets, and COLD~\cite{deng2022cold}, ToxiCN~\cite{toxicn}, SWSR~\cite{jiang2022swsr}, and CDial-Bias~\cite{zhou2022towards}, which are Chinese toxicity datasets. The OffensEval 2019, AbusEval, and HatEval datasets are derived from Twitter and focus on offensive language, abusive language, and hate speech, respectively. The RAL-E dataset, sourced from a banned Reddit community, is a large-scale, unannotated English dataset. In contrast, ToxiGen is a toxicity dataset generated using GPT-3, targeting multiple groups. The COLD, SWSR, CDial-Bias, and ToxiCN datasets are collected from Chinese social media platforms, including Zhihu, Weibo, and Tieba, each focusing on different groups. Compared to these datasets, ours features the largest collection of toxicity data, with each text annotated with a toxicity score, providing researchers with a valuable resource to optimize and evaluate the safety of LLMs.


\begin{table}[htbp!]
\begin{center} 
    \begin{threeparttable}
        \caption{Comparison of Different Toxicity Datasets.}
        \label{tab:comparison_toxic}
        \begin{tabular}{*{8}{c}}
            \toprule 
            Dataset & Lang. & Label & Toxicity Score & Data Size & Toxic & Benign \\ 
            \midrule
            OffensEval 2019~\cite{offenseval} & EN & \textcolor{green}{\usym{2714}} & \textcolor{red}{\usym{1F5F4}} & 14,100 & 4,640 & 9,460 & \\
            AbusEval~\cite{abuseval} & EN & \textcolor{green}{\usym{2714}} & \textcolor{red}{\usym{1F5F4}} & 14,100 & 2,927 & 11,173 & \\
            HatEval~\cite{hateval} & EN & \textcolor{green}{\usym{2714}} & \textcolor{red}{\usym{1F5F4}} & 13,000 & 5,470 & 7,530 & \\
            RAL-E~\cite{hatebert} & EN & \textcolor{red}{\usym{1F5F4}} & \textcolor{red}{\usym{1F5F4}} & 1,492,740 & - & - & \\
            ToxiGen~\cite{hartvigsen2022toxigen} & EN & \textcolor{green}{\usym{2714}} & \textcolor{red}{\usym{1F5F4}} & 274,186 & 137,395 & 136,791 & \\
            COLD~\cite{deng2022cold} & ZH & \textcolor{green}{\usym{2714}} & \textcolor{red}{\usym{1F5F4}} & 37,480 & 18,041 & 19,439 & \\
            ToxiCN~\cite{toxicn} & ZH & \textcolor{green}{\usym{2714}} & \textcolor{red}{\usym{1F5F4}} & 12,011 & 6,461 & 5,550 & \\
            SWSR~\cite{jiang2022swsr} & ZH & \textcolor{green}{\usym{2714}} & \textcolor{red}{\usym{1F5F4}} & 8,969 & 3,093 & 5,876 & \\
            CDial-Bias~\cite{zhou2022towards} & ZH & \textcolor{green}{\usym{2714}} & \textcolor{red}{\usym{1F5F4}} & 28,343 & 7,233 & 21,110 & \\
            \textbf{ChineseWebText2.0} & ZH & \textcolor{green}{\usym{2714}} & \textcolor{green}{\usym{2714}} & 1,803,836,772 & 1,632,620 & 1,802,204,152 & \\
            \bottomrule
        \end{tabular}
    \end{threeparttable}
\end{center}
\end{table}

\section{Conclusions and Future Work}
\label{others}


In order to extract large-scale, high-quality Chinese pre-training data with multi-dimensional and fine-grained information, we have developed a novel pipeline that processes raw collected data using both handcrafted rules and advanced evaluation models. The handcrafted rules are applied to clean the raw texts, producing a high-quality dataset. Subsequently, we design three powerful models—a quality evaluation, a domain classifier, and a toxicity evaluation model—to assign each text with four types of fine-grained information. With this approach, we release the latest and largest Chinese dataset of 3.8 TB, each of which is annotated with a quality score, domain labels, a toxicity score and a toxicity label. This dataset enables LLM researchers to re-filter the data according to their desired thresholds. Additionally, We also release the complete tool-chain that transforms the raw data into a high-quality dataset with multi-dimensional and fine-grained annotations.

In the future, we will collect a wider variety of domain-specific texts to enable the dataset to cover a broader range of scenarios. Additionally, we are going to enrich the dataset by incorporating more fine-grained information into the texts. For example, we could construct a evaluation model to assess the knowledge density within each text.

\bibliographystyle{unsrt}  
\bibliography{references}  

\section*{Appendix}
\subsection{Human Evaluation in Quality Assessment}
\label{appendix:human_evaluation}
Following the methodology used in ChineseWebText 1.0, we employ human evaluators to assess the quality of the dataset. In this process, we randomly sample 1,000 examples, and five independent human evaluators are hired to assess the quality of each sample. The evaluation is based on four key criteria. Each text is assigned a label of either "True" or "False" during the evaluation. "True" indicates that the data meets the quality requirements for pre-training across all four criteria, while "False" indicates that the text contains some level of noise. After the evaluations are completed, we calculate the average accuracy across the five evaluators.
\begin{tcolorbox}[colback=lightblue!50!white,colframe=lightblue,title=\textcolor{black}{The Criteria for Human Evaluation in Quality Assessment.}, width=\textwidth]

\begin{itemize}
    \item Informativeness: Whether the text contains enough knowledge and information, or is just meaningless crap.
    \item Fluency: Whether the text has formatting issues, capitalization mistakes, or evident grammatical errors that impair readability.
    \item Coherence: Whether the text progressively forms a coherent body of information on a topic through its successive sentences.
    \item Toxicity: Texts used for pre-training should endeavor to exclude offensive remarks, sexually explicit content, and politically sensitive statements to mitigate potential generative risks.
\end{itemize}

\end{tcolorbox}

\end{document}